\documentclass[pdflatex,sn-mathphys-num]{sn-jnl}

\usepackage{graphicx}%
\usepackage{multirow}%
\usepackage{amsmath,amssymb,amsfonts}%
\usepackage{amsthm}%
\usepackage{mathrsfs}%
\usepackage[title]{appendix}%
\usepackage{xcolor}%
\usepackage{textcomp}%
\usepackage{manyfoot}%
\usepackage{booktabs}%
\usepackage{algorithm}%
\usepackage{algorithmicx}%
\usepackage{algpseudocode}%
\usepackage{listings}%

\theoremstyle{thmstyleone}%
\theoremstyle{thmstyletwo}%

\theoremstyle{thmstylethree}%

\begin{document}

\title[Article Title]{
PolyCrysDiff: Controllable Generation of Three-Dimensional Computable Polycrystalline Material Structures
}


\author[1]{\fnm{Chi} \sur{Chen}}\email{chi-chen@sjtu.edu.cn}
\author[1]{\fnm{Tianle} \sur{Jiang}}\email{jiangtianle4@sjtu.edu.cn}
\author[1]{\fnm{Xiaodong} \sur{Wei}}\email{xiaodong.wei@sjtu.edu.cn}
\author*[2]{\fnm{Yanming} \sur{Wang}}\email{yanming.wang@sjtu.edu.cn }

\affil[1]{\orgdiv{Global College}, \orgname{Shanghai Jiao Tong University}, \orgaddress{\street{No.800 Dong Chuan Road}, \city{Shanghai}, \postcode{200240}, \state{Shanghai}, \country{China}}}

\affil[2]{\orgdiv{Global Institute of Future Technology}, \orgname{Shanghai Jiao Tong University}, \orgaddress{\street{No.800 Dong Chuan Road}, \city{Shanghai}, \postcode{200240}, \state{Shanghai}, \country{China}}}


\abstract{The three-dimensional (3D) microstructures of polycrystalline materials exert a critical influence on their mechanical and physical properties. Realistic, controllable construction of these microstructures is a key step toward elucidating structure-property relationships, yet remains a formidable challenge. Herein, we propose PolyCrysDiff, a framework based on conditional latent diffusion that enables the end-to-end generation of computable 3D polycrystalline microstructures. Comprehensive qualitative and quantitative evaluations demonstrate that PolyCrysDiff faithfully reproduces target grain morphologies, orientation distributions, and 3D spatial correlations, while achieving an $R^2$ over 0.972 on grain attributes (e.g., size and sphericity) control, thereby outperforming mainstream approaches such as Markov random field (MRF)- and convolutional neural network (CNN)-based methods. The computability and physical validity of the generated microstructures are verified through a series of crystal plasticity finite element method (CPFEM) simulations. Leveraging PolyCrysDiff’s controllable generative capability, we systematically elucidate how grain-level microstructural characteristics affect the mechanical properties of polycrystalline materials.  This development is expected to pave a key step toward accelerated, data-driven optimization and design of polycrystalline materials.}

\keywords{Polycrystalline material, 3D Diffusion, Material generation, CPFEM}



\maketitle
\newpage
\section{INTRODUCTION}\label{sec1}

The microstructure of polycrystalline materials exerts a decisive influence on their physical and mechanical properties~\cite{Hall-Petch, Grain-boundary, Grain-orientation}. Accordingly, understanding and optimizing  microstructural features is central to the design of high-performance polycrystalline materials. Recent advances in computational modeling and machine learning offer promising pathways to accelerate this process~\cite{Optimal-polyhedral, active-learning, opt-alloy, Opt-battery-material}. By integrating high-throughput simulations with data-driven optimization strategies, the discovery of polycrystalline materials with unprecedented properties can be significantly expedited. In this context, accurate modeling of three-dimensional (3D) polycrystalline microstructures has become increasingly important. There is a growing demand for methods capable of efficiently generating high-quality, physically realistic, controllable, and computatable microstructures.


Unlike molecular or atomic systems, polycrystalline materials exhibit intrinsically more complex structures and larger spatial scales, making their modeling considerably more challenging, particularly when computational feasibility is considered. First, grain orientation must be explicitly represented to enable subsequent simulations. The model should accurately capture the crystallographic orientations of all grains without redundancy or loss of information, while maintaining a physically consistent spatial distribution throughout the structure. In addition, the generated structures must exhibit clear and well-defined grains while avoiding diffuse or unrealistic grain boundaries, which imposes stringent requirements on structural quality.

Experimental techniques such as serial sectioning~\cite{Serial-sectioning} and X-ray computed tomography (CT)~\cite{CT} can reconstruct real 3D microstructures; however, these methods are both costly and time-consuming, rendering them impractical for large-scale structural optimization. Consequently, numerical modeling approaches have been developed to generate synthetic polycrystalline structures. Deterministic geometry-based methods, such as Neper~\cite{Neper} and Dream.3D~\cite{DREAM3D}, commonly employ algorithms based on voronoi~\cite{Voronoi} or laguerre voronoi tessellations to construct grain structures with controllable parameters. While these methods are computationally efficient and widely used for general microstructure modeling, their representational flexibility remains limited. The strict geometric constraints inherent to these algorithms restrict their ability to reproduce irregular morphologies and complex spatial correlations observed in real materials. Furthermore, their conditional generation often relies on iterative loops, leading to potential issues with convergence and computational cost.

Machine learning (ML) methods have been explored for polycrystalline microstructure modeling due to their powerful pattern recognition and generative capabilities. Texture synthesis, a classical computer vision technique designed to produce visually similar textures, has been adapted to model polycrystalline materials, which can be viewed as textured patterns composed of grains with distinct orientations. A representative example is the Markov Random Field (MRF) method, originally developed for generating two-dimensional (2D) texture images based on local neighborhood similarity between pixels~\cite{MRF_2D}. Sundararaghavan et al. extended this method to 3D by comparing local voxel neighborhoods with exemplar 2D micrographs to reconstruct 3D polycrystalline structures~\cite{MRF_3D}, with the underlying formulation provided in Eqs. (S1–S3).. Although MRF-based methods can reconstruct 3D microstructures from a limited number of 2D images, they are computationally expensive due to voxel-level comparisons and rely on empirically tuned parameters, often resulting in suboptimal quality. Subsequent works introduced perceptual loss functions, such as the VGG-based loss~\cite{VGG}, to mitigate the limitations of pixel-wise matching and improve both computational efficiency and perceptual realism~\cite{VGGTexture2D}. Building on this concept, several neural network–based models have been proposed for volumetric texture generation. For instance, SolidTextureNet employs convolutional neural networks (CNNs) to synthesize 3D textures by minimizing the VGG loss between slices of the generated volume and reference images~\cite{SolidTexture}, while GramGAN utilizes generative adversarial networks (GANs) for 3D texture generation~\cite{GramGAN}. Bostanabad et al. formulated voxel generation as an optimization problem using perceptual loss functions~\cite{PixelOpt}. Overall, these texture synthesis–based approaches demonstrate the potential of ML for polycrystalline modeling and can reconstruct 3D structures from exemplar images. However, since they do not explicitly learn 3D spatial dependencies, the resulting structures often lack sufficient physical realism and fidelity.

Recent advances in deep learning—particularly in generative models—have made high-quality synthesis increasingly feasible, owing to their ability to learn underlying data distributions from large-scale datasets. Diffusion models~\cite{DDPM} have achieved state-of-the-art performance in image and volumetric data generation, while Stable Diffusion further introduces a powerful conditioning mechanism that enables precise control over generated outputs~\cite{StableDiffusion}. Their effectiveness has also been demonstrated across various material-related applications~\cite{MolecularDiffusion, ProteinDiffusion, mf-diffusion, PorousDiffusion}. Together, these advantages make diffusion models a highly promising method for conditional generation of high-quality, computable 3D polycrystalline material structures.

In this work, we propose PolyCrysDiff, a framework based on conditional latent diffusion for  generation of computable 3D polycrystalline material structures. The model employs a 3D variational autoencoder (VAE) to learn a compact latent representation of microstructures and utilizes a diffusion process to generate high-quality structures under the guide of conditional inputs. To evaluate the performance of our model, we first train an unconditional version to assess its capability for random generation. Subsequently, a conditional version is introduced to demonstrate controllable generation under specific property constraints, where two representative conditional inputs—the mean grain size (implemented as the grain number within a fixed volume) and the mean grain sphericity—are employed. Furthermore, an application-oriented workflow is established, integrating feature specification, microstructure generation, and subsequent crystal plasticity finite element method (CPFEM) simulations~\cite{CPFEM} to assess material properties. This workflow not only verifies the computational feasibility of the generated structures but also demonstrates the potential of PolyCrysDiff for data-driven material optimization and design. Together, these methods enable a systematic evaluation of the geometric, crystallographic, and mechanical consistency of the generated microstructures. The proposed PolyCrysDiff framework provides a robust and extensible approach for high-quality, controllable, and computable 3D polycrystalline structure generation, paving the way for accelerated microstructure optimization and materials innovation.

\section{RESULTS}\label{sec2_results}

\subsection{Model architecture}
\begin{figure}[htbp]
    \centering
    \includegraphics[width=0.95\linewidth]{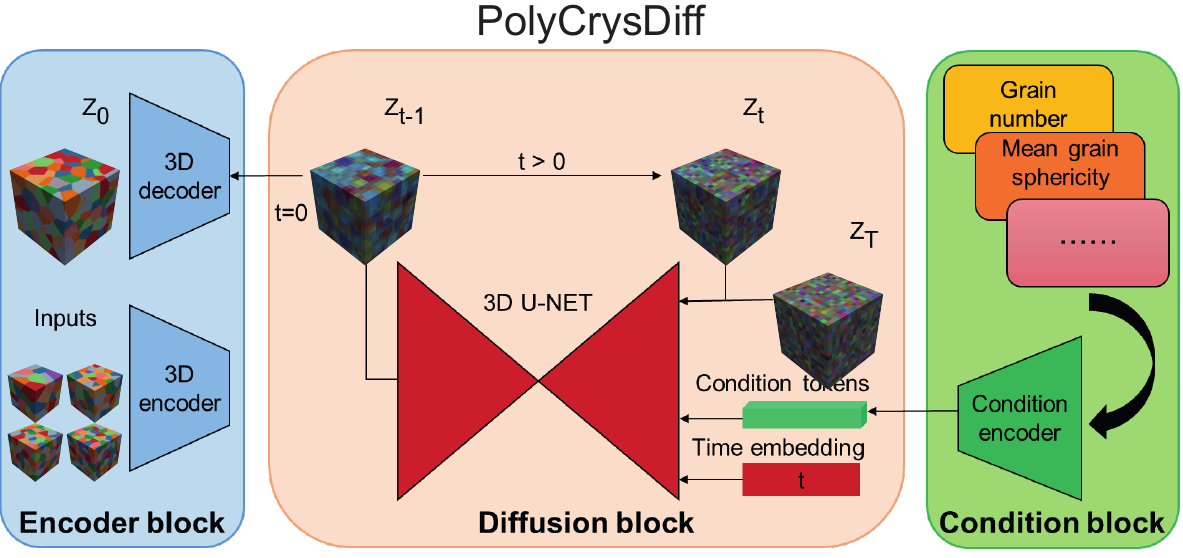}
    \caption{Model architecture of the proposed PolyCrysDiff framework. The blue blocks represent the encoder–decoder components of the 3D variational autoencoder (VAE). The green blocks denote the conditional generation module.}
    \label{fig:model_architecture}
\end{figure}

In this work, polycrystalline structures are represented as voxel-based volumes with dimensions of 
$(64,64,64)$. Each voxel is assigned an RGB value corresponding to its crystallographic orientation, analogous to an electron backscatter diffraction (EBSD) image. In this representation, microstructural features such as grain boundaries and orientation gradients are naturally preserved.

The architecture of the proposed PolyCrysDiff model is illustrated in Fig. \ref{fig:model_architecture}. It follows the general design of Stable Diffusion model~\cite{StableDiffusion}, extended to 3D by replace all the convolution layers with 3D convolutions. A 3D VAE is employed to map high-dimensional polycrystalline structures into a compact latent space, with its performance presented in Fig. S1, thereby reducing computational complexity and facilitating stable model training. The diffusion process is then performed within this latent space.During training, a clean polycrystalline structure is encode into the latent space. Its latent vector is progressively perturbed with Gaussian noise according to a fixed noise schedule. A 3D U-Net serves as the denoising network, trained to predict the added noise at each timestep. In the reverse(generation) process, sampling begins from random Gaussian noise, which is iteratively denoised using the trained U-Net to reconstruct a clean latent representation. The decoder of the VAE then transforms this latent sample bach into voxel space, yielding a generated 3D polycrystalline microstructure.The diffusion algorithm can be summarize by  equation \ref{equation_DDPM}~\cite{DDPM} and a detailed visualization of the generation procedure is provided in Fig. S2.

\begin{equation}
    \mathbf{x}_{t-1} = \frac{1}{\sqrt{\alpha_t}} \left( \mathbf{x}_t - \frac{1 - \alpha_t}{\sqrt{1 - \bar{\alpha}_t}} \boldsymbol{\epsilon}_\theta(\mathbf{x}_t, t) \right) + \sigma_t \mathbf{z}
    \label{equation_DDPM}
\end{equation}

To enable controllable generation, a conditioning module is incorporated into the network. Conditioning inputs—such as class labels, physical parameters are first embedded by a condition encoder and then injected into the diffusion process via cross-attention layers within the U-Net. This mechanism allows the model to generate microstructures consistent with specified conditions while maintaining structural diversity.

\subsection{Unconditional generation}
\begin{figure}[htbp]
    \centering
    \includegraphics[width=0.95\linewidth]{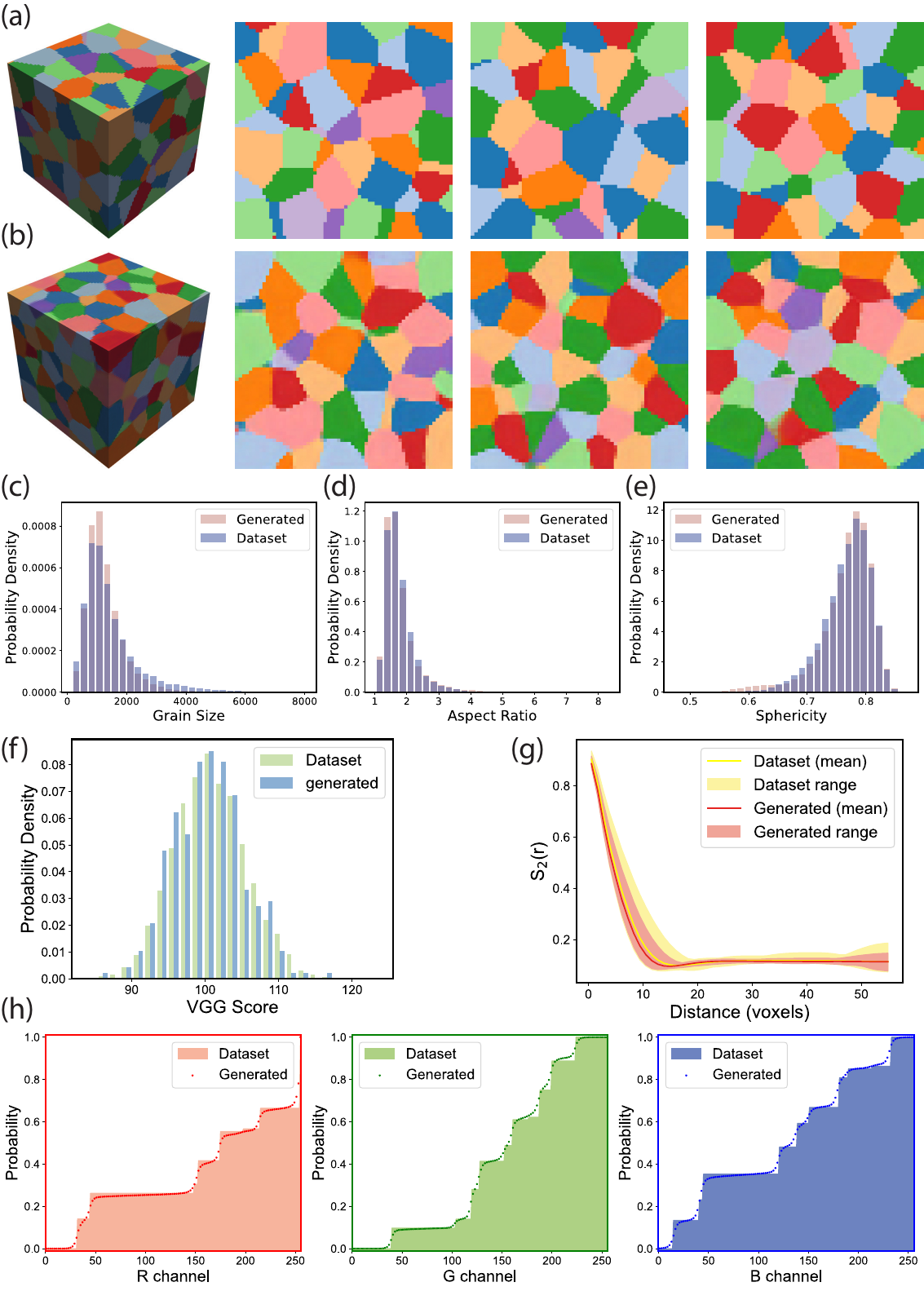}
    \caption{Qualitative and quantitative evaluation of the proposed PolyCrysDiff model for unconditioned
    polycrystalline microstructure generation (a) 3D view and orthogonal slices of a reference polycrystalline microstructure from the training dataset. (b) 3D view and slices of an unconditionally generated microstructure with a comparable grain count to (a).  (c)-(e) Distributions of grain size, aspect ratio, and sphericity for generated and reference structures. (f) Distribution of VGG perceptual similarity scores for 2D slices sampled from generated and reference 3D microstructures.(g) Range and mean of the two-point correlation function ($S_2$) for both datasets. (h) Cumulative distribution functions of RGB channels representing crystallographic orientation.}
    
    \label{fig:uncondition}
\end{figure}

The unconditional diffusion model was first trained and evaluated with the conditional block disabled. In this setting, the model learns the intrinsic statistical distribution of the training dataset and generates polycrystalline material structures that follow this distribution. The generation process is entirely stochastic, as it starts from random noise during inference with no external conditions applied.
A total of 100 structures were generated and analyzed.

Fig. \ref{fig:uncondition} provides a comprehensive evaluation of the proposed generative model from both qualitative and quantitative perspectives. In Fig. \ref{fig:uncondition} (a) and (b),  we present a reference polycrystalline microstructure from the dataset  alongside a microstructure generated by our PolyCrysDiff model with a comparable grain count for fair visual comparison. For each model, a 3D rendering of the entire microstructure and representative 2D slices along the x-, y-, and z- axes are shown, allowing assessment of both global morphology and local grain features. In general, the generated microstructure closely replicates the characteristics of the reference model while also exhibiting structural novelty. The strong consistence of geometric features and color distribution demonstrated that the model effectively captures the underlying grain topology and orientation statistics. Moreover, the continuity of grains across orthogonal slices indicates that the generative approach preserves 3D consistency rather than merely producing visually plausible but uncorrelated slices. These results collectively demonstrate the model’s capability to generate realistic 3D polycrystalline microstructures.

To further evaluate the consistency and quality of the generated structures, quantitative analyses were performed from both grain-level and voxel-level perspectives. Each structure from the generated set and training dataset was segmented to obtain grains separately, and the grain properties were then calculated; more details are provided in \ref{sec:structure_analysis}. In Fig. \ref{fig:uncondition} (c-e), the histograms of grain size, aspect ratio, and sphericity show excellent agreement in both shape and peak position between the generated and reference microstructures. This visual consistency is quantitatively supported by the small Kolmogorov–Smirnov(KS) (0.0357–0.0858) and Earth Mover’s Distance(EMD) (0.0051–0.0176) values reported in Table \ref{table:uncondition}, indicating that the distributions of all three grain properties are closely matched. The strong consistency between the generated and dataset distributions confirms that the model successfully learn the grain characteristic from training dataset, including shape anisotropy, grain morphology and size variability.

VGG score of slices from 3D structures are investigated to indicate perceptual similarity between them. For each 3D structure, we randomly select three slices from each axis respectively to ensure that the slices can effectively represent the 3D structure. Histogram of VGG scores from generated set and training set is shown in Fig. \ref{fig:uncondition} (f), while the corresponding statistical metrics are given in Table~\ref{table:uncondition}, with low values of $KS=0.0280$ and $EMD=0.0085$. This highlight the model’s ability to generate microstructures that are not only statistically consistent but also perceptually indistinguishable from real data. The two-point correlation function $S_2(r)$~\cite{S2} was also calculated for each structure to evaluate and compare their spatial correlations. For each set, the range and mean value of their curves are presented in Fig. \ref{fig:uncondition} (g). The very close curves show the neighborhood environment is very well preserved in the generated structures. Finally, the cumulative distribution functions (CDFs) of the RGB channels shown in Fig. \ref{fig:uncondition} (h) confirm that the generated voxel orientations faithfully follow the crystallographic orientation distribution of the dataset. 

\begin{table}[htbp]
\centering
\caption{Statistical comparison between generated and reference microstructures.}
\label{table:uncondition}
\begingroup
\setlength{\tabcolsep}{12pt} 
\begin{tabular}{lcccc}
\toprule
Metric & Grain size & Aspect ratio & Sphericity & VGG score \\
\midrule
KS    & 0.0662    & 0.0858       & 0.0357     & 0.0280 \\
EMD   & 0.0155    & 0.0176       & 0.0051     & 0.0085 \\
\bottomrule
\end{tabular}
\endgroup
\end{table}

Together, these results validate that the proposed model successfully captures both the geometric and crystallographic characteristics of the polycrystalline microstructures in the training set, demonstrating its potential for downstream simulations and materials design tasks. Moreover, the high fidelity of the generated structures highlights the generalizability of our approach. Since the training dataset is based on Voronoi-generated microstructures without any material-specific features, the model’s performance suggests strong potential for extension to other classes of polycrystalline materials. With sufficient experimental or simulation data, the same framework can be readily adapted to generate realistic 3D structures with different morphological characteristics, further broadening its applicability beyond the current dataset.

\subsection{Generation with controlled parameters}

\begin{figure}[htbp]
    \centering
    \includegraphics[width=0.95\linewidth]{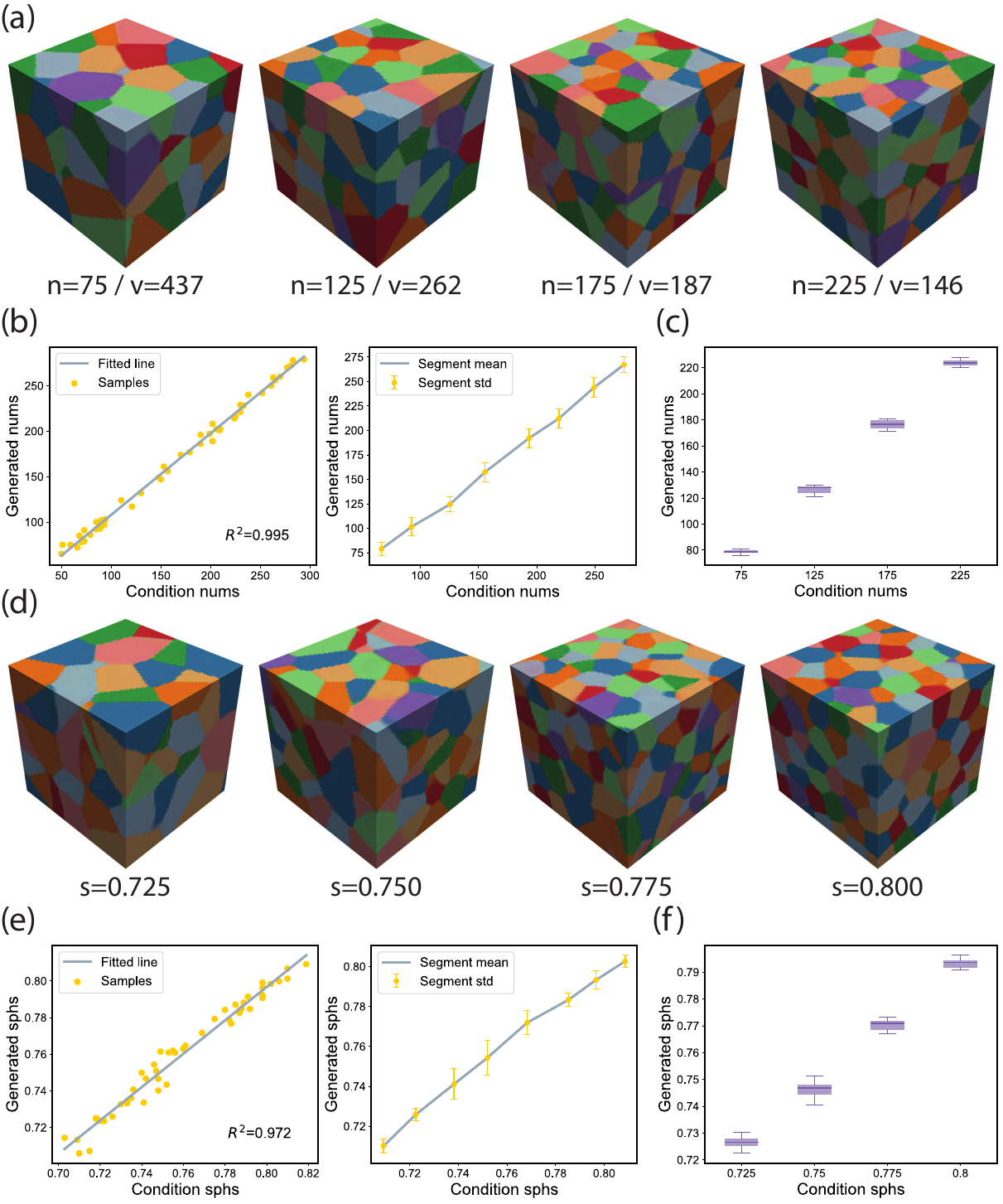}
    \caption{Qualitative and quantitative evaluation of the proposed PolyCrysDiff model for conditioned polycrystalline microstructure generation. (a) Microstructures generated under mean grain size conditions of 437, 262, 187, and 146 \textmu m$^3$, v stands for mean grain size(\textmu m$^3$) and n stands for total grain number. (b) Correlation between target and generated grain numbers. (c) Distribution of generated grain numbers for six samples per condition, demonstrating stability and repeatability. (d) Microstructures generated under mean sphericity conditions of 0.725, 0.75, 0.775, and 0.800, s stands for grain sphericity. (e) Correlation between target and generated sphericity values for grain mean sphericity. (f) Variability of generated sphericity across repeated samples for each condition.}
    \label{fig:condition}
\end{figure}

To enhance the controllability and practical applicability of our generative framework, we extend it to conditional generation, enabling the synthesis of microstructures that satisfy user-specified target properties rather than producing purely random samples. Such controllability is crucial in polycrystalline material design, where microstructures with prescribed grain characteristics are often required to explore structure–property relationships and guide materials optimization.  In the proposed framework, conditioning is implemented by introducing condition tokens, which is encoded from physical property values, into the diffusion process via cross-attention layers. Linear layers are used as encoder here. 

A class-label-conditioned model is trained, with its results shown in Fig. S3. And in the following, we demonstrate the controllbility of our model through two representative value-conditioned generation tasks: (1) controlling the mean grain size and (2) controlling the mean grain sphericity.  For the mean grain size control task, implementation is by controlling the total grian numbers within the fixed volume. These tasks showcase the model’s ability to generate 3D polycrystalline microstructures with tunable morphological characteristics.

Fig. \ref{fig:condition} summarizes the results of the conditional generation experiments. Fig. \ref{fig:condition} (a) illustrates visual results for the mean grian size control task. Four structures generated under conditional inputs of 75, 125, 175, and 225 grains, respectively, clearly exhibit an decreasing mean grain size from left to right, demonstrating the model’s intuitive control capability. To quantitatively assess this performance, 50 structures were generated with grain number condition values randomly sampled between 50 and 300 (consistent with the training data range), while sixteen representative generated 3D structures are shown in Fig. S4 to further illustrate the performance of the conditional model. The relationship between the target and the actual grain number is plotted in Fig. \ref{fig:condition} (b), showing a strong linear correlation ($R^2 = 0.995$). A binned mean–standard error plot further confirms stable performance across the entire range. Model robustness was also evaluated by generating six independent samples for each of four condition grain numbers (75, 125, 175, 225). As shown in Fig. \ref{fig:condition} (c), the resulting grain numbers are tightly distributed around their target values, demonstrating stable and repeatable control. Fig. \ref{fig:condition} (d–f) present analogous results for the mean grain sphericity condition. In Fig. \ref{fig:condition} (d), four microstructures generated under target sphericity values of 0.725, 0.75, 0.775, and 0.800 exhibit an evident morphological transition from sharper, faceted grains to smoother, more equiaxed shapes as the sphericity increases. Quantitative results shown in Fig. \ref{fig:condition} (e) and (f) confirm the high accuracy and stability of conditional control, with generated values closely following the target trend.

\subsection{Comparison with existing models} 
\begin{figure}[htbp]
    \centering
    \includegraphics[width=0.95\linewidth]{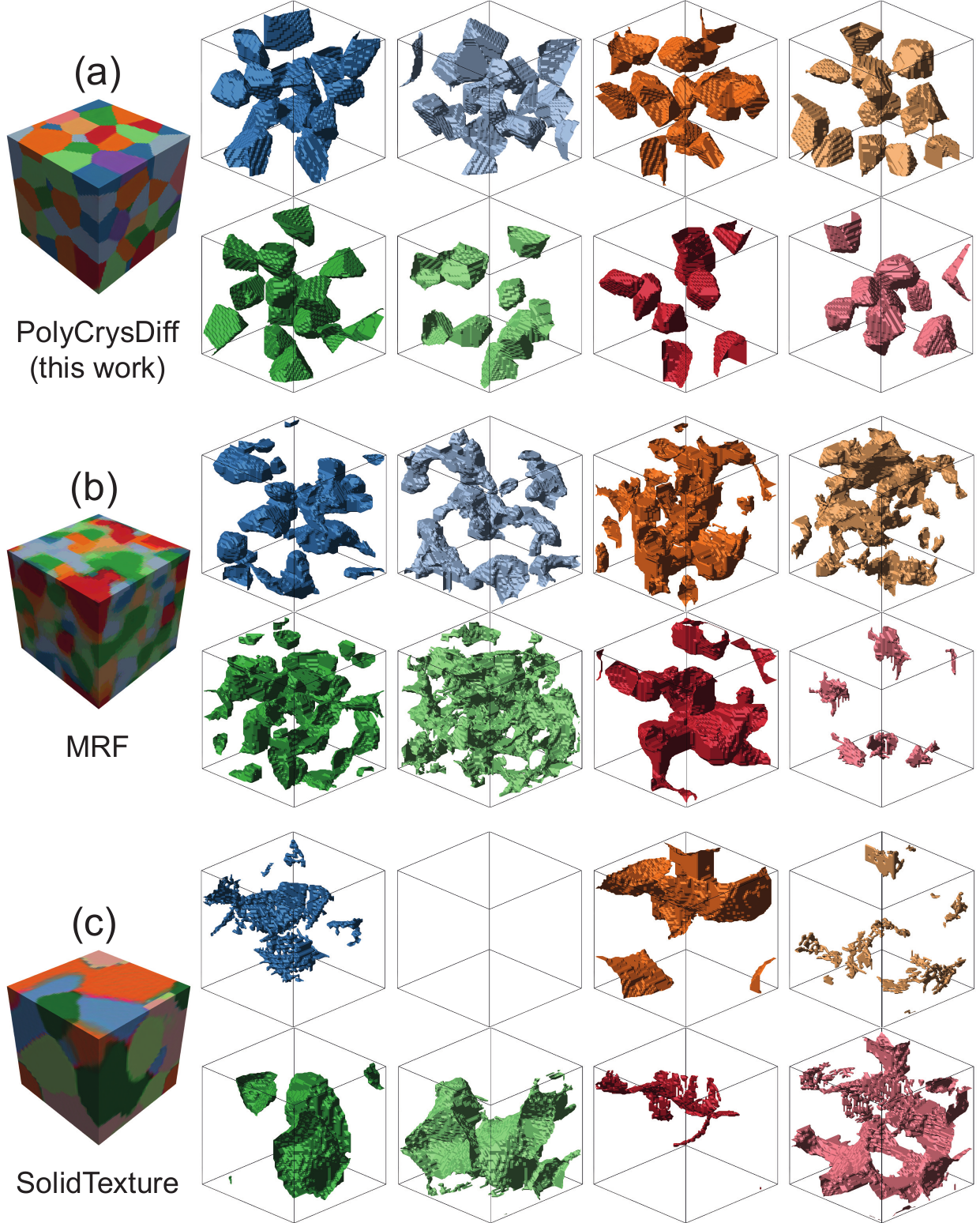}
    \caption{(a) Microstructure generated by the proposed PolyCrysDiff model, showing clear grain morphology, sharp grain boundaries, and consistent orientation distribution.
    (b) Microstructure generated by the MRF model, having indistinguishable grains and unrealistic structures.
    (c) Microstructure generated by the SolidTexture model, which fails to reconstruct coherent 3D grain structures and exhibits missing orientations.}
    \label{fig:comparison}
\end{figure}

To further demonstrate the superiority of our model in generating high-quality and computationally feasible polycrystalline structures under conditional generation, we select a 3D microstructure generated by PolyCrysDiff and use its slice as the input for both the MRF model and the SolidTexture model for comparison. The resulting grain structures from the three methods are visualized and compared in Fig. \ref{fig:comparison}, with additional results from the MRF and SolidTexture models provided in Fig. S5 and S6, respectively.

Fig. \ref{fig:comparison} (a) shows the structure generated by PolyCrysDiff, where each grain exhibits a well-defined morphology with sharp, clean grain boundaries and a physically consistent orientation distribution. In contrast, the MRF-generated structure in Fig. \ref{fig:comparison} (b) fails to produce distinct grains—voxels of identical orientation are chaotically connected, making it impossible to differentiate individual grains. This results in unrealistic microstructures that cannot be used for further grain-level analysis or computational simulations. Similarly, the SolidTexture model output shown in Fig. \ref{fig:comparison} (c) also fails to reconstruct coherent 3D grain structures. Moreover, certain orientations are missing (the second orientation), indicating a loss of crystallographic information and further compromising the structural validity.

These results highlight that our framework achieves superior conditional generation performance, producing physically plausible 3D microstructures with clear grain morphology and accurate orientation distributions. Such high-quality and computable structures provide a reliable foundation for subsequent materials simulation and design.

\subsection{End to end generation and CPFEM results}

\begin{figure}[htbp]
    \centering
    \includegraphics[width=0.95\linewidth]{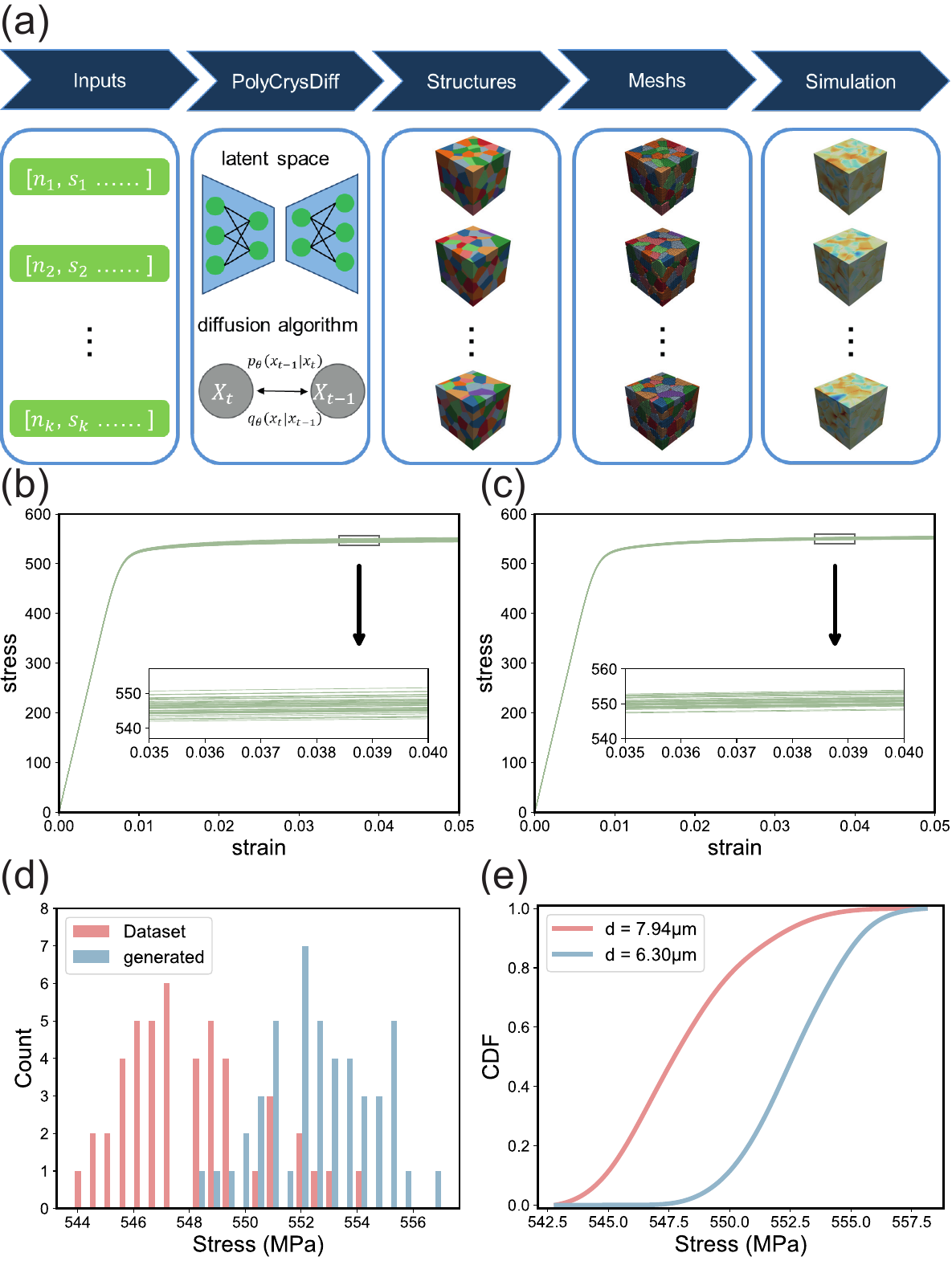}
    \caption{(a) Workflow of the proposed end-to-end framework based on PolyCrysDiff. User-defined condition inputs are given to PolyCrysDiff and structures satisfying these inputs are generated. Structures can be easily meshed and then CPFEM simulation is applied. (b) Stress–strain curves from CPFEM simulations of 50 PolyCrysDiff-generated structures conditioned on a grain count of 125. (c) Stress–strain curves from CPFEM simulations of 50 PolyCrysDiff-generated structures conditioned on a grain count of 250. (d) Histogram of the tensile strength distribution for results in (b) and (c). Grain count are represented in mean grain size. (e) PDFs of the CPFEM-predicted tensile strength.}
    \label{fig:framework}
\end{figure}

Beyond the model itself, we further propose an end-to-end framework that leverages the computability and controllability of the generated structures to support material design and structural optimization. The controllability of the PolyCrysDiff model enables users to efficiently generate polycrystalline structures with target parameters, thereby facilitating systematic exploration of the design space. Meanwhile, the computability of the generated structures allows seamless integration with simulation tools for evaluating their physical behavior.

Here, we demonstrate this capability through CPFEM simulations, where the generated structures are subjected to uniaxial tensile loading to obtain their mechanical responses and material properties. The overall workflow is illustrated in Fig. \ref{fig:framework}(a). Specifically, user-defined structural parameters such as grain number or sphericity distribution are provided as conditional inputs to the PolyCrysDiff model, which generates corresponding 3D polycrystalline structures. After meshing these structures, CPFEM simulations are conducted using the MOOSE framework to perform tensile tests. Meshed structures and corresponding 3D stress distributions are provided here. After obtaining simulation results for a series of generated structures, their mechanical properties—such as Young’s modulus and tensile strength—can be quantitatively analyzed. By correlating these properties with the corresponding conditional inputs, the structure–property relationships can be systematically investigated. 

We performed a case study to examine the dependence of tensile strength on the mean grain size d in polycrystalline materials. Two sets of microstructures were generated by conditioning the model on the mean grain size d=7.94 $\mu \text{m}$ (group 1) and d=6.30 $\mu \text{m}$ (group 2). Each group contains 50 samples. CPFEM simulations were conducted for all microstructures. All stress–strain curves for the two sets are shown in Fig. \ref{fig:framework}(b) and (c). The tensile strength of each sample was extracted from the stress–strain response, and the statistical distributions are summarized in Fig. \ref{fig:framework}(d) and (e). The mean tensile strengths are 548.0MPa for group 1 and 552.6MPa for group 2, while the whole range is only (543.9, 557.1), showing a clear difference between two groups. In our experiments, samples with a smaller mean grain size exhibit higher tensile strength. This process provides valuable insights for microstructure optimization and accelerates data-driven materials design.

\section{Discussion}\label{sec4-Conclusion}
In this work, we introduced PolyCrysDiff, a conditional latent diffusion model capable of generating realistic and simulation-ready 3D polycrystalline microstructures. By combining a VAE-based latent representation with diffusion and cross-attention conditioning, the framework achieves high-fidelity reconstruction of complex grain morphologies while enabling accurate and stable control over key microstructural attributes such as mean grain size (total grain count) and sphericity. Comparisons with existing models demonstrate the improvements brought by our approach. The controllability of PolyCrysDiff is particularly valuable for exploring structure–property relationships. In our grain-size case study, CPFEM simulations on PolyCrysDiff-generated samples revealed a clear strengthening trend for smaller grains, consistent with established experimental understanding. This highlights the practical potential of the framework for accelerated microstructure studies and data-driven materials design.

Several opportunities remain for future development. Incorporating real 3D data from experiments may improve realism and broaden the diversity of generated structures. For conditioning inputs, vectorized descriptors could offer finer control, and more complex modalities such as images or text descriptions may further enhance flexibility. Overall, PolyCrysDiff provides a powerful and efficient approach for generating controllable 3D grain structures and establishes a promising foundation for microstructure engineering and polycrystalline materials optimization.

\section{METHODS}\label{se3c_method}

\subsection{Training details}
The polycrystalline microstructures used in this study were generated using a 3D Voronoi tessellation algorithm~\cite{Voronoi}. Each training dataset contains 2,000 samples, represented as voxel grids with dimensions of $(64, 64, 64, 3)$, where the last channels correspond to RGB values encoding crystallographic orientations. All voxel values are normalized to the range $[-1, 1]$. The number of grains in each sample is randomly sampled between 50 and 300, and seeded within the domain with a regularity factor of 0.5~\cite{regularity}.

For grain orientations (colors), ten fixed RGB values are selected to represent discrete crystallographic directions. A greedy coloring algorithm is employed to assign colors such that adjacent grains have distinct orientations whenever possible. Geometric descriptors including grain sphericity, mean grain size, and other statistical metrics are simultaneously computed for each structure, providing the conditional inputs used during conditional model training.

3D VAE is first trained to encode the high-dimensional microstructures into a latent space of size $(16, 16, 16)$. Subsequently, the diffusion model is trained in this latent domain. Both models are optimized using the Adam optimizer with an initial learning rate of $1 \times 10^{-6}$. The batch size is set to 2 for VAE training and 1 for the diffusion model. A linear noise schedule with $T = 1000$ timesteps is adopted, where $\beta_t$ increases from $1.5 \times 10^{-3}$ to $0.02$.

All models are implemented in PyTorch and trained on a single NVIDIA RTX 4090 GPU using mixed-precision computation for improved efficiency. Training is performed for 400 epochs with a cosine learning rate scheduler and gradient clipping to stabilize convergence. The total training time is approximately 19 hours for the VAE and 70 hours for the diffusion model. A mean squared error (MSE) loss is employed in the latent space to measure the reconstruction quality during training.

\subsection{Structure analysis}\label{sec:structure_analysis}

\begin{figure}[htbp]
    \centering
    \includegraphics[width=0.95\linewidth]{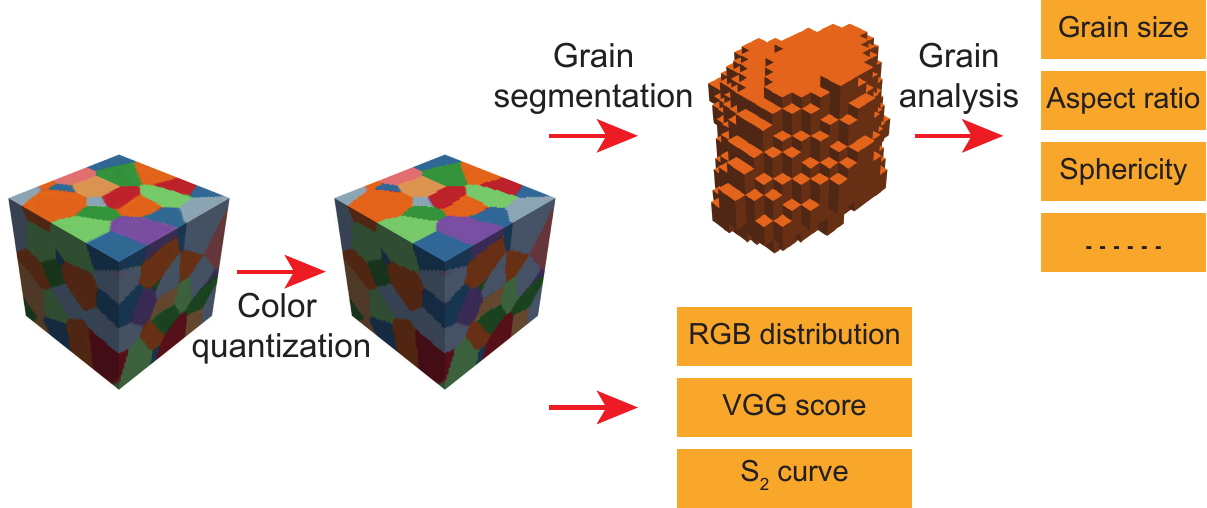}
    \caption{Polycrystalline material structure analysis process.}
    \label{fig:model_analysis}
\end{figure}
 
To quantitatively evaluate and analyze the 3D polycrystalline structures from both the training dataset and the generated dataset, several statistical and geometric metrics are employed. These metrics are divided into two main categories: voxel-level descriptors and grain-level descriptors.
The overall analysis workflow is illustrated in Fig. \ref{fig:model_analysis}. For each given structure, we first perform color quantization, in which the color of each voxel is mapped to the nearest representative orientation color. This step simplifies subsequent quantitative analysis and ensures consistent orientation representation. After quantization, the RGB value distribution within the volume is directly obtained. For the VGG-based perceptual score, 2D slices are required. In our experiments, one slice is randomly selected from each principal axis to provide a representative view of the 3D structure. Then, for each orientation, a corresponding binary mask is constructed, and its two-point correlation function $S_2(r)$ is computed using an efficient fast fourier transform (FFT)–based autocorrelation method. The overall $S_2(r)$ curve of the structure is obtained as the weighted average of all orientation-specific results.

For the grain-level analysis, individual grains are identified through connectivity-based segmentation. Very small grains that arise from discretization noise are removed using a size threshold, which is set to 200 voxels in our experiments. To further separate adjacent or partially merged grains, the watershed algorithm~\cite{Watershed} is applied. Once segmentation is complete, the voxel set of each grain is available, from which grain volume and principal axis lengths can be computed directly. The grain surface area is estimated using the marching cubes algorithm~\cite{marching_cube}, which reconstructs a smooth boundary surface. Based on these geometric quantities, additional morphological descriptors such as sphericity and aspect ratio are calculated.

\subsection{Distribution similarity metrics}
The similarity between the generated and reference microstructures is quantified using the KS distance \cite{Massey1951} and the normalized EMD \cite{Rubner2000}. Both metrics are computed from the empirical distributions of grain size, aspect ratio, and sphericity.

The KS distance measures the maximum deviation between two empirical CDFs:
\begin{equation}
D_{\mathrm{KS}} = \sup_x \left|F_g(x) - F_r(x)\right|.
\end{equation}

To complement this local discrepancy measure, the EMD evaluates the minimal transport cost required to transform one distribution into another. For empirical distributions $p$ and $q$, the EMD corresponds to the 1-Wasserstein distance:
\begin{equation}
W_1(p, q) = \inf_{\gamma \in \Pi(p, q)} \sum_{i,j} \gamma_{ij}\, d_{ij},
\end{equation}
where $\Pi(p, q)$ denotes valid couplings and $d_{ij}$ is the ground distance between histogram bins. To ensure comparability across grain properties with different physical scales, we report the normalized EMD:
\begin{equation}
W_{1,\mathrm{norm}} = \frac{W_1(p, q)}{x_{\max} - x_{\min}}.
\end{equation}

Together, $D_{\mathrm{KS}}$ captures the maximum pointwise deviation between distributions, while $W_{1,\mathrm{norm}}$ reflects the global transport cost, providing a complementary assessment of distributional similarity.

\subsection{CPFEM simulation}\label{s4_CPFEM}
In this work, all CPFEM simulations are performed using the MOOSE framework~\cite{MOOSE}, with the governing equations provided in Eqs. (S4–S7). The material modeled is Al7075-T6. A total engineering strain of 5\% was applied along the z-axis and the simulation was advanced to a total simulated time of 5 s. The cubic elastic stiffness constants for Al7075-T6 are set to $C_{11} = 107.3$ GPa, $C_{12}=60.9$ GPa and $C_{44}=28.3$ GPa~\cite{Al7075parameters1,Al7075parameters2}. A complete list of model parameters is provided in Table S1.

\subsection{Framework test case}
The grain size $d$ is defined by formulation \ref{d_formula}, where $V$ is the total volume and n is the number of grains. In our case, d=7.94$\mu \text{m}$ means 125 grains in the total volume and d=6.30$\mu \text{m}$ is 250 grains. The tensile strength is characterized by the ultimate stress at 5\% applied strain. 
For the CDF curve, kernel density estimation (KDE) method is used to get a smooth probability density function (PDF) and then accumulated to get it.

\begin{equation}
    d = (\frac{6V}{n\pi})^{\frac{1}{3}}
    \label{d_formula}
\end{equation}








\newpage
\bibliography{sn-bibliography}

\section{Data availability}
The authors declare that the data supporting the findings of this study are available within the paper and its Supplementary Information files. Should any raw data files be needed in another format they are available from the corresponding author upon reasonable request.

\section{Acknowledgement}
The computations in this paper were run on the Siyuan-1 and Zhiyuan-1 clusters supported by the Center for High Performance Computing at Shanghai Jiao Tong University.

\section{Fundings}
This work was supported by Shanghai Science and Technology Commission Program (25DZ3001902).

\section{Author Contributions}
C.C. developed the PolyCrysDiff framework, generated datasets, performed model training and validation, conducted data analysis and visualization, and drafted the manuscript. T.J. carried out crystal plasticity finite element method (CPFEM) simulations and analyzed the mechanical responses of both generated and reference 3D polycrystalline microstructures. X.W. provided guidance on the research and contributed to scientific discussions. Y.W. supervised the research and provided overall project guidance. All authors reviewed and edited the manuscript.

\section{Conflict of Interest}
The authors declare that they have no competing interests as defined by Nature Portfolio, or other interests that might be perceived to influence the results and/or discussion reported in this paper.

\end{document}